\documentclass[10pt,twocolumn,letterpaper]{article}

\usepackage[pagenumbers]{cvpr} 

\definecolor{navyblue}{HTML}{0071BC}
\definecolor{hotpink}{HTML}{FF0080}

\definecolor{oai-white}{HTML}{FFFFFF}
\definecolor{oai-black}{HTML}{000000}
\definecolor{oai-red}{HTML}{FF4500}
\definecolor{oai-green}{HTML}{51DA4C}
\definecolor{oai-blue}{HTML}{0000FF}
\definecolor{oai-yellow}{HTML}{FFF639}
\definecolor{oai-magenta}{HTML}{FF45FF}
\definecolor{oai-cyan}{HTML}{00FFFF}
\definecolor{oai-orange}{HTML}{FE7600}
\definecolor{oai-violet}{HTML}{8A2BE2}
\definecolor{oai-brown}{HTML}{A0522D}
\definecolor{oai-green-050}{HTML}{F4FFF4}
\definecolor{oai-green-100}{HTML}{E9FFE8}
\definecolor{oai-green-200}{HTML}{D9FFD8}
\definecolor{oai-green-300}{HTML}{C9FFC7}
\definecolor{oai-green-400}{HTML}{A6FFA3}
\definecolor{oai-green-500}{HTML}{7CF178}
\definecolor{oai-green-600}{HTML}{51DA4C}
\definecolor{oai-green-700}{HTML}{3FA93B}
\definecolor{oai-green-800}{HTML}{2D712A}
\definecolor{oai-green-900}{HTML}{193718}
\definecolor{oai-gray-000}{HTML}{FFFFFF}
\definecolor{oai-gray-100}{HTML}{FAFAFA}
\definecolor{oai-gray-200}{HTML}{F5F5F5}
\definecolor{oai-gray-300}{HTML}{E5E5E5}
\definecolor{oai-gray-400}{HTML}{FFB7A4}
\definecolor{oai-gray-500}{HTML}{CDCDCD}
\definecolor{oai-gray-600}{HTML}{A8A8A8}
\definecolor{oai-gray-700}{HTML}{747474}
\definecolor{oai-gray-800}{HTML}{393939}
\definecolor{oai-gray-900}{HTML}{000000}

\definecolor{eclipseBlue}{RGB}{42,0.0,255}
\definecolor{eclipseGreen}{RGB}{63,127,95}
\definecolor{eclipsePurple}{RGB}{127,0,85}
\definecolor{lightgray}{rgb}{0.98,0.98,0.98}

\definecolor{codeBlue}{RGB}{0,0,180}
\definecolor{codeGreen}{RGB}{0,128,0}
\definecolor{codePurple}{RGB}{128,0,128}
\definecolor{bgGray}{RGB}{245,245,245}

\definecolor{mybg}{HTML}{F3FAFD}
\definecolor{myframe}{HTML}{41535B}
\definecolor{mymagenta}{HTML}{a02b93}
\definecolor{visual}{HTML}{A50E0E}       
\definecolor{linguistic}{HTML}{174EA6}   
\definecolor{relational}{HTML}{E37400}   
\definecolor{egocentric}{HTML}{0D652D}  
\definecolor{cvprblue}{rgb}{0.21,0.49,0.74}
\usepackage[pagebackref,breaklinks,colorlinks,allcolors=cvprblue]{hyperref}
\usepackage[accsupp]{axessibility}
\usepackage{colortbl}
\usepackage{xcolor}
\usepackage{rotating}
\usepackage{multirow}
\usepackage{graphicx}
\usepackage{pifont}
\usepackage[accsupp]{axessibility}
\usepackage{listings}
\usepackage{enumitem}        
\usepackage{fontawesome5}
\usepackage[most]{tcolorbox}
\tcbuselibrary{skins, breakable}

\def\paperID{} 
\def\confName{CVPR}
\def\confYear{2026}

\title{From Where Things Are to What They Are For: \\ Benchmarking   Spatial–Functional Intelligence in Multimodal LLMs}

 \newcommand{\finding}[2]{
      \vspace{-0.05cm}
      \begin{tcolorbox}[
          enhanced,
          colback=teal!4!white,
          colframe=teal!60!black,
          arc=2pt,
          boxrule=0pt,
          leftrule=3pt,
          toprule=0pt,
          bottomrule=0pt,
          rightrule=0pt,
          boxsep=2pt,
          left=8pt,
          right=6pt,
          top=2pt,
          bottom=2pt,
          fontupper=\small,
          before upper={\textbf{\textcolor{teal!70!black}{~Finding~#1.}}\;\;},
      ]
      #2
      \end{tcolorbox}
      \vspace{-0.25cm}
  }

\newcommand{\fmtnum}[1]{%
  \ifnum\fpeval{#1 < 0} = 1
    \textcolor{red}{$#1$}%
  \else
      \textcolor{green}{$#1$}%
  \fi
}

\definecolor{linkblue}{rgb}{0.1, 0.5, 0.7}

\lstdefinelanguage{jsonCompact}{
    basicstyle=\ttfamily\footnotesize,
    commentstyle=\color{codeGreen},
    stringstyle=\color{codeBlue},
    numbers=left,
    numbersep=4pt,
    numberstyle=\tiny\color{gray},
    breaklines=true,
    showstringspaces=false,
    upquote=true,
    string=[s]{"}{"},
    morecomment=[l]{:"},
    literate=
     *{:}{{{\color{black}{:}}}}{1}
      {,}{{{\color{black}{,}}}}{1}
      {\{}{{{\color{black}{\{}}}}{1}
      {\}}{{{\color{black}{\}}}}}{1}
      {[}{{{\color{black}{[}}}}{1}
      {]}{{{\color{black}{]}}}}{1},
}

\newtcolorbox{promptbox}[2][]{
    enhanced,
    attach boxed title to top left={yshift=-3mm, xshift=3mm}, 
    colback=gray!5,              
    colframe=gray!60!black,      
    colbacktitle=gray!70!black,  
    coltitle=white,              
    title={#2},                  
    fonttitle=\bfseries\small\sffamily,
    fontupper=\small\sffamily,   
    boxrule=0.5pt,
    arc=2pt,
    #1                           
}

\lstdefinelanguage{promptJson}{
    basicstyle=\ttfamily\scriptsize,
    keywords={true,false,null},
    keywordstyle=\color{blue}\bfseries,
    stringstyle=\color{brown!60!black},
    commentstyle=\color{green!50!black},
    breaklines=true,
    columns=fullflexible,
    frame=single,               
    framerule=0.5pt,            
    rulecolor=\color{gray!30},  
    backgroundcolor=\color{white}, 
    aboveskip=5pt,
    belowskip=5pt
}

\newcommand{\huggingface}{\raisebox{-1.5pt}{\includegraphics[height=1.05em]{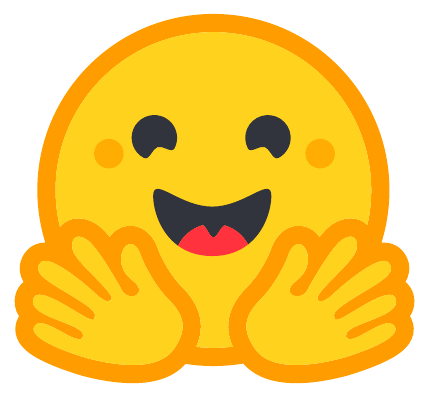}}\xspace}
\author{
      Le Zhang\textsuperscript{1}\protect
  \quad
      Jihan Yang\textsuperscript{2} \quad
      Soundarya Krishnan\textsuperscript{3}  \quad
      Jimit Majmudar\textsuperscript{3}  \quad  \\
      Xiou Ge\textsuperscript{3} \quad
      Prasoon Puri\textsuperscript{3}  \quad
      Prathamesh Saraf\textsuperscript{3}  \quad
      Shruti Bhargava\textsuperscript{3}  \quad
      Dhivya Piraviperumal\textsuperscript{3} \quad \\
      Yinan Ling\textsuperscript{3} \quad
      Cindy Pan\textsuperscript{3}  \quad
      Hong Yu\textsuperscript{3}  \quad
      Aishwarya Agrawal\textsuperscript{1,4} \quad
      Bo-Hsiang Tseng\textsuperscript{3} \quad \\ \\
      \textsuperscript{1}Mila - Qu\'{e}bec AI Institute, UdeM \quad
      \textsuperscript{2}NYU \quad 
      \textsuperscript{3}Apple \quad 
      \textsuperscript{4}Canada CIFAR AI Chair \quad \\ [0.5em]
     \href{https://github.com/lezhang7/spatial-function-reasoning}{\textcolor{black}{\faGithub}~Code} \quad \quad 
      \href{https://lezhang7.github.io/sfi-bench.github.io/}{\textcolor{black}{\faGlobe}~Website} \quad \quad
      {\huggingface \href{https://huggingface.co/datasets/le723z/SFI-Bench}{{\text{SFI-Bench}}}}
\vspace{0.2cm}
  }

\begin{document}

\twocolumn[{
    \renewcommand\twocolumn[1][]{#1}
    \maketitle
    \centering
    \captionsetup{type=figure}
    \includegraphics[width=1\textwidth]{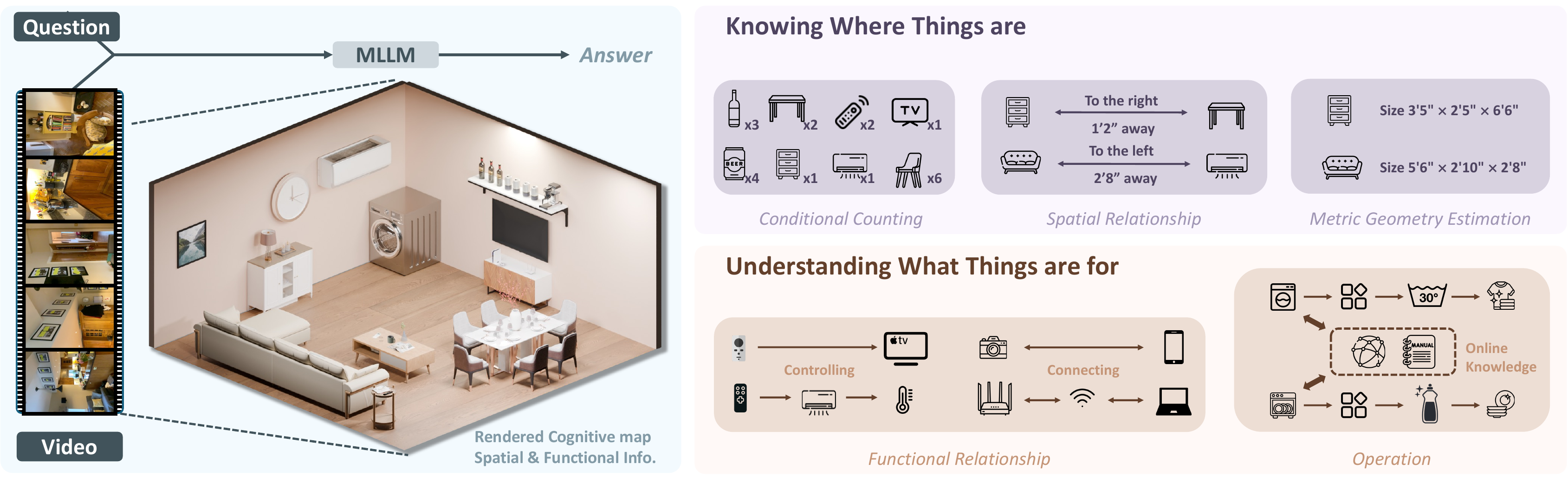}
   \caption{\textbf{From Spatial Cognition to Intelligent Agents.}
\textbf{Left:} Task pipeline. A video is provided as input, and a multimodal model must reason across frames over both spatial and temporal context to answer a question.
\textbf{Right:} Two complementary reasoning abilities evaluated in our benchmark.
\textbf{Top (\textit{Where Things Are})}: spatial reasoning that requires understanding the scene layout and geometric relationships among objects (e.g., counting, directions, distances, and size).
\textbf{Bottom (\textit{What They Are For})}: functional reasoning that requires understanding object affordances and functional relationships within the environment, enabling goal-oriented action and planning.
}
    \label{fig:teaser}
\vspace{8mm}
}]

\renewcommand{\thefootnote}{\fnsymbol{footnote}}

\begin{abstract}

Human-level agentic intelligence extends beyond low-level geometric perception, evolving from recognizing where things are to understanding what they are for. While existing benchmarks effectively evaluate the geometric perception capabilities of multimodal large language models (MLLMs), they fall short of probing the higher-order cognitive abilities required for grounded intelligence. To address this gap, we introduce the Spatial–Functional Intelligence Benchmark (SFI-Bench), a video-based benchmark with over 1,500 expert-annotated questions derived from diverse egocentric indoor video scans. SFI-Bench systematically evaluates two complementary dimensions of advanced reasoning: (1) Structured Spatial Reasoning, which requires understanding complex layouts and forming coherent spatial representations, and (2) Functional Reasoning, which involves inferring object affordances and their context-dependent utility. The benchmark includes tasks such as conditional counting, multi-hop relational reasoning, functional pairing, and knowledge-grounded troubleshooting, directly challenging models to integrate perception, memory, and inference. Our experiments reveal that current MLLMs consistently struggle to combine spatial memory with functional reasoning and external knowledge, highlighting a critical bottleneck in achieving grounded intelligence. SFI-Bench therefore provides a diagnostic tool for measuring progress toward more cognitively capable and truly grounded multimodal agents.

\end{abstract}

\section{Introduction}

Humans navigate and interact with their environment by forming internal \textit{cognitive maps}—structured representations that capture both the spatial organization of objects and their potential uses. 
These representations support a range of behaviors, from spatial reasoning to goal-directed interaction with tools. 
For artificial agents, approximating this capability requires going beyond visual recognition to infer two complementary forms of structure: 
a \textbf{spatial representation} that captures object layouts and relational structure, and a \textbf{functional representation} that encodes affordances\footnote{Our notion of \emph{functional reasoning} is related to the concept of \emph{affordance} in psychology~\cite{gibson2014ecological}, which describes the action possibilities that the environment offers to an agent. In this work, we adopt a narrower, operational view, focusing on object--function associations (e.g., whether an object can support a given use), rather than modeling the full agent-dependent and perceptual aspects of affordance.}, purposes, and context-dependent usage.

Recent advances in Multimodal Large Language Models (MLLMs)~\cite{bai2025qwen2,comanici2025gemini,gpt4o,gpt5,glm,yang2025qwen3} have brought us closer to this goal, powering modern vision--language--action (VLA) systems~\cite{kim2024openvla,black2024pi_0,cheang2024gr,cheang2025gr}. 
Yet, systematically evaluating whether these models truly acquire such spatial and functional intelligence remains challenging~\cite{black2024pi_0,atreya2025roboarena}. 
Existing benchmarks such as VSI-Bench~\cite{yang2025thinking} primarily probe the \emph{first} step of this developmental hierarchy—testing geometric perception and factual recall—while leaving the higher cognitive stages of structured map construction, affordance inference, and knowledge-grounded reasoning largely unexamined.

To close this gap, we introduce the \textbf{Spatial--Functional Intelligence Benchmark (SFI-Bench)}, which holistically evaluates cognitive intelligence across progressive stages (see \cref{fig:teaser}). 
While prior works contain tasks labeled as counting or spatial relations, these are typically formulated as \textit{perceptual} recognition problems.
SFI-Bench instead reformulates them as \textit{cognition-level} challenges: conditional counting demands logical and compositional inference over attributes and relations (\textit{e.g., finding the maximum number of same-brand bottles on a cabinet}) and path reasoning requires integrating spatial cues across multiple views over time to infer a coherent global layout. These tasks incentivize models to build coherent internal representations of space rather than merely reacting to local cues.

Beyond spatial cognition, SFI-Bench incorporates functional and knowledge-grounded reasoning, probing whether models understand what objects in the scene are for, how they are operated, and how failures can be diagnosed.
Tasks such as functional pairing, operational planning, and causal troubleshooting assess whether a model can bridge perception to action, mirroring the functional reasoning that underlies human goal-directed behavior. 
This shifts the evaluation from testing spatial memory to evaluating the broader pre-action cognitive abilities required for agentic behaviour.

Evaluating state-of-the-art MLLMs on SFI-Bench reveals a consistent pattern: while current models excel at local perception, they remain brittle in maintaining global spatial memory, grounding affordances, and composing multi-step functional plans. 
Our analyses uncover several key findings. 
First, longer reasoning chains do not lead to better decisions; reasoning quality saturates once a moderate budget is reached, beyond which overthinking introduces semantic drift. 
Second, cognitive map construction depends strongly on visual evidence rather than textual descriptions, and—unlike humans—models exhibit surprising insensitivity to temporal continuity. 
Third, systematic failure modes arise across tasks, including visual ambiguity, object recognition errors, spatial layout inconsistencies, and affordance overgeneralization.

In addition, SFI-Bench reveals the crucial role of \textit{external knowledge acquisition}. 
For operational and troubleshooting tasks, GPT-5 exhibits a performance gap of up to 8\%, depending on whether web search is enabled, highlighting that many functional questions fundamentally require grounding in up-to-date or device-specific knowledge. 
This underscores an often-overlooked challenge for multimodal reasoning systems \cite{hong2025deepeyesv2, narayan2025deepmmsearch}: the need to seamlessly integrate visual perception with dynamic, external knowledge sources rather than relying on closed-world parametric memory alone.

Together, these findings point to a critical frontier for multimodal AI: moving beyond perception-oriented models toward systems capable of integrated spatial–functional cognition—constructing, maintaining, and exploiting coherent cognitive maps while flexibly retrieving and applying external knowledge to support purposeful action in real-world environments.

\begin{figure*}[!thb]
    \centering
    \includegraphics[width=\linewidth]{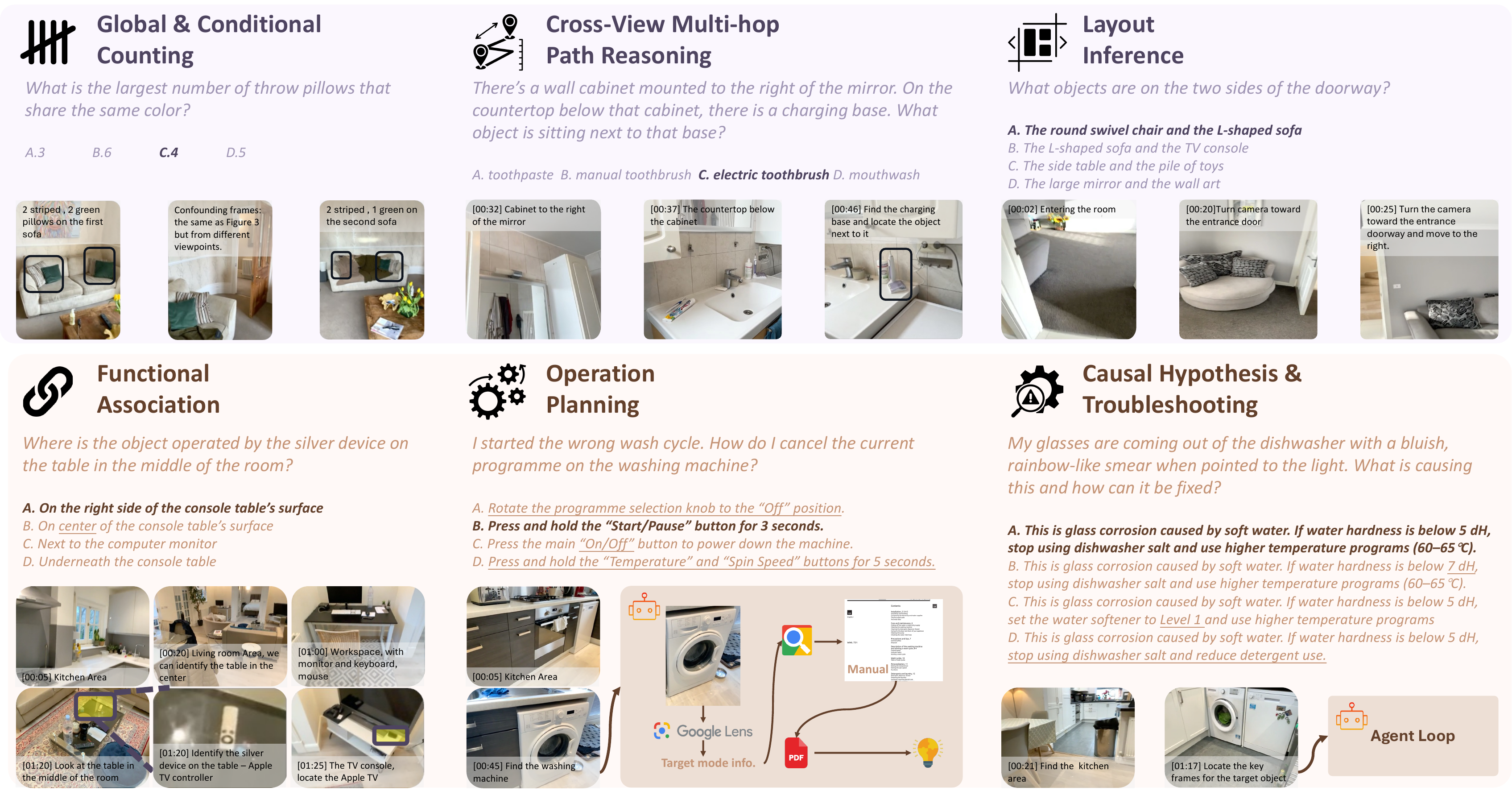}
\caption{\textbf{Task examples in SFI-Bench.} Required grounding cues and reasoning evidence are highlighted using bounding boxes and accompanying text. For functional reasoning tasks, potentially confounding options are underlined. Answers are simplified for readability. For the final two functional reasoning tasks, successful completion requires online search to access relevant operational manuals.}
    \label{fig:overview}
\end{figure*}
\section{SFI-Bench}

\subsection{Dataset Overview}

We introduce \textbf{SFI-Bench} to evaluate how multimodal foundation models acquire cognitive abilities for intelligent agents. SFI-Bench is a \emph{video-based multiple-choice question answering} benchmark with 1555 \emph{human-annotated} questions from 134 real-world egocentric indoor videos, sourced from ARKitScenes \cite{dehghan2021arkitscenes} and ScanNet++ \cite{yeshwanth2023scannet++}, covering diverse spatial layouts and functional contexts in residential, professional, and industrial environments. SFI-Bench spans six core tasks (illustrated in \cref{fig:overview}) grouped into two fundamental cognitive capabilities central to agentic intelligence:

\subsubsection{Cognitive Spatial Reasoning.}  These tasks assess whether a model can move beyond frame-level perception to construct \emph{structured cognitive spatial maps}.
Rather than recognizing objects in isolation, the model must compositionally integrate attributes, absolute and relative positions, and multi-view spatial cues distributed across the video.
This requires stitching together fragmented observations to form a coherent and temporally consistent representation.

    \textbf{Global and Conditional Counting.}  Reformulates counting as a \emph{compositional $\&$ logical reasoning} task.  Beyond simple enumeration, models must apply attribute constraints and perform set-based operations—such as intersection, union, and complement—along with group-level aggregation  (e.g., identifying the largest subset of same-brand bottles on a cabinet).  This shifts counting from perceptual detection to structured logical inference.

    \textbf{Cross-View Multi-hop Path Reasoning.}  
    Evaluates the ability to integrate spatial evidence across time and viewpoints to infer relationships not visible in any single frame.  
    Success requires constructing a coherent multi-hop spatial memory and recovering implicit connections between objects and locations beyond immediate perception.

    \textbf{Layout Inference.}  
    Evaluates whether the model can integrate distributed cues into a coherent global scene layout and reason about \emph{occlusion relationships}. 
Because referenced objects often never appear together, the model must infer their relative arrangement and visibility ordering across frames. 
This reflects real-world navigation, where understanding occlusions is essential for building a consistent spatial map.

\subsubsection{Functional Reasoning.}  These tasks evaluate whether a model can move from spatial understanding to \emph{functional cognition}—inferring object affordances, interactions, and context-dependent use.  
These tasks require integrating visual evidence with \textit{external knowledge sources} (e.g., device manuals, online instructions), testing whether models can retrieve, interpret, and apply functional knowledge in real-world scenarios.

    \textbf{Functional Association.}  
    Tests whether the model can infer \emph{affordance relationships} between objects.  
    Objects often never co-occur in the same frame; thus, the model must link them through cues such as brand, design, or spatial context  
    (e.g., associating a remote with the correct television), reflecting early functional map construction.

    \textbf{Operation Planning.}  
    Probes whether a model can determine \emph{how an object should be used}.  
    Solving these questions requires searching for device-specific information (e.g., manuals), interpreting retrieved knowledge, and assembling multi-step action plans grounded in the videos.

    \textbf{Causal Hypothesis and Troubleshooting.}  
    Assesses a model’s ability to diagnose problems by combining scene understanding with external knowledge. The model must hypothesize plausible failure modes, consult relevant documentation via \emph{web search}, and integrate the two sources to generate a grounded and actionable solution.

\begin{figure*}[!htb]
    \centering
    \includegraphics[width=\linewidth]{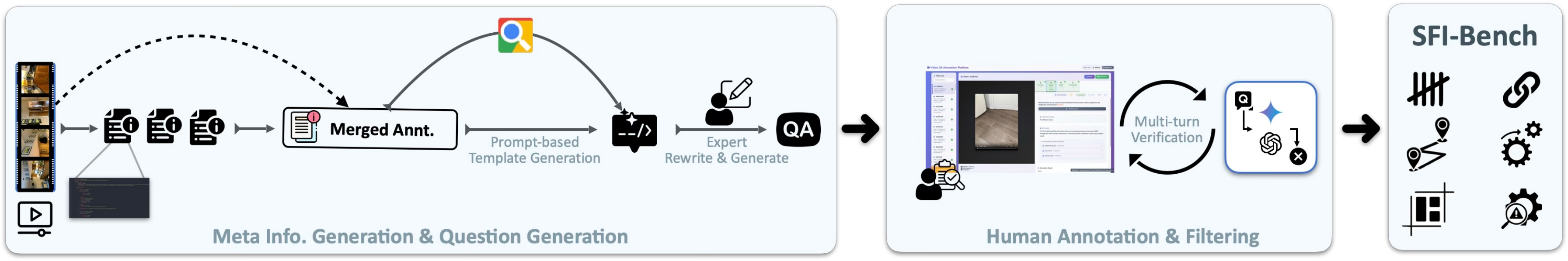}
\caption{\textbf{Benchmark curation pipeline.}
Metadata is extracted and consolidated across multiple MLLM passes, and combined with task-specific templates and few-shot examples to generate candidate questions. Annotators verify all questions and provide answers for the first four tasks, while answers for the two knowledge-grounded tasks are derived from expert-retrieved manuals. Finally, all questions undergo multi-turn human--AI collaborative post-hoc filtering to ensure quality and consistency. Samples that models fail to answer are manually rechecked, and questions that can be solved without visual grounding are removed.}

    \label{fig:curation pipeline}

\end{figure*}

\begin{figure}[htb]
    \centering
    \includegraphics[width=\linewidth]{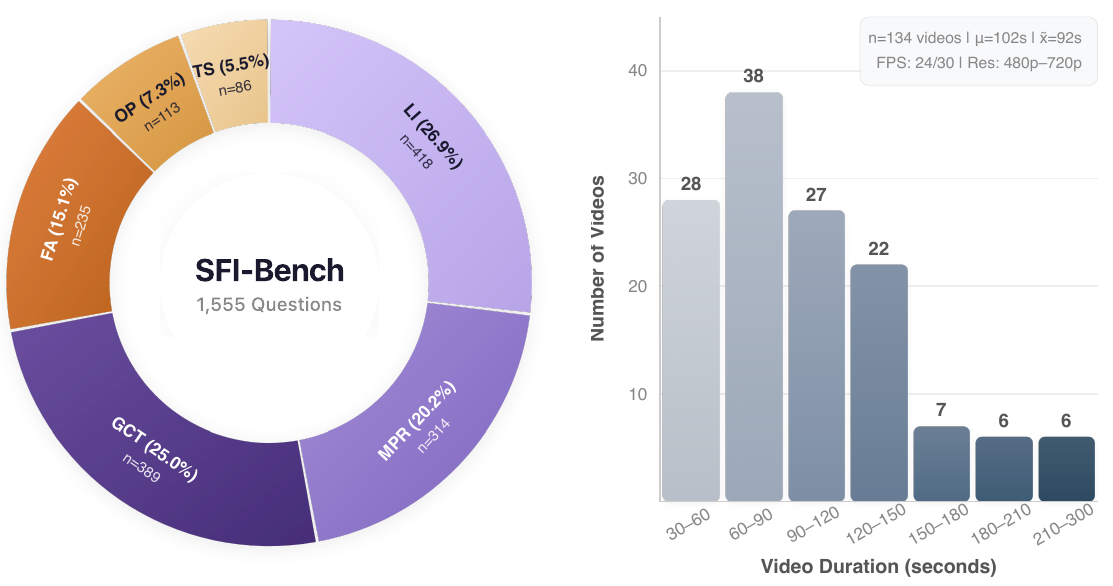}
    \caption{\textbf{Dataset Statistics.} Task and video length distribution. Task names are abbreviated for brevity.}
    \label{fig:Statistics}
  
\end{figure}

\subsection{Benchmark Curation Process}

SFI-Bench is constructed through a three-stage pipeline designed to produce high-quality, temporally grounded questions (details in Appendix~\ref{app:app_datacurationdetails}; pipeline shown in \cref{fig:curation pipeline}).

\noindent\textbf{Automatic Question Generation.}
We repurpose egocentric scans from ARKitScenes and ScanNet++ and use Gemini-2.5-Pro to extract fine-grained metadata for each video.
Multiple passes of metadata extraction are merged and cross-validated against the raw video to obtain a reliable structured description of objects, attributes, spatial relations, and functional roles.
Task-specific templates then generate candidate questions; for knowledge-grounded tasks, relevant manuals are retrieved online by annotators manually and integrated into the generation prompts.
Annotators \emph{refine all generated questions}, correct mismatches, and add additional items that better capture each task’s cognitive requirements. The statistics are illustrated in \cref{fig:Statistics}.

\noindent\textbf{Human Verification and Answer Annotation.}
For the first four tasks, annotators watch each video, validate every question, and provide ground-truth answers based on visual evidence. For the two knowledge-grounded tasks, answers are automatically generated from retrieved device manuals.

\noindent\textbf{Post-hoc Quality Filtering.}
All questions undergo automated and manual validation. Each item is first evaluated using Gemini-2.5 Pro and GPT-5, and any incorrectly answered case is then reviewed through a \emph{multi-turn human--AI verification process} to diagnose potential issues and refine the question or options when necessary. Questions that can be solved without videos are removed to ensure visual dependence.
\section{Benchmarking on SFI-Bench}
\subsection{Evaluation Setups}

\vspace{0.05cm}
\noindent\textbf{Baseline Models.} We comprehensively evaluate a wide range of MLLMs \emph{capable of processing video inputs}, spanning both open-source and proprietary systems. Among proprietary systems, we benchmark Gemini-3, Gemini-2.5 ~\cite{comanici2025gemini}, GPT-5.4/5~\cite{gpt5}, and o4-mini under default configurations. 
Open-source models include Qwen3-VL~\cite{yang2025qwen3}, InternVL-3.5~\cite{zhu2025internvl3}, GLM-4.5~\cite{hong2025glm}, LLaVA-OneVision~\cite{li2024llavanext-ablations}, and LLaVA-Video~\cite{llavavideo}.
All evaluations are conducted in a zero-shot setting using same prompt templates to ensure fairness and reproducibility.

\vspace{0.05cm}
\noindent\textbf{Evaluation Metric.} All samples in SFI-Bench are multiple-choice questions (MCQ), each with four candidate options (25\% random chance). Performance is measured by answer accuracy. 
For the first four tasks—\textit{Conditional Counting}, \textit{Path Reasoning}, \textit{Layout Inference}, and \textit{Functional Association}—models are directly prompted to select the correct option. 
For the remaining tasks—\textit{Functional Planning} and \textit{Causal Hypothesis \& Troubleshooting}—models equipped with search tools are allowed to retrieve external knowledge (e.g., user manuals) via web search before answering. Models without tool-use or web-access capabilities are evaluated in the same offline setting as the first four tasks.

\begin{figure*}[ht!]
    \captionsetup{type=table}
    \vspace{-0.4cm}
    \centering
    \begin{minipage}{0.63\textwidth} 
    \centering
    \fontsize{4.6pt}{4.4pt}\selectfont
    \setlength\tabcolsep{3pt} 
    \renewcommand{\arraystretch}{1.2} 
    \scalebox{1.57}{
    \begin{tabular}{r|cc|cccccc}
Methods & Rank & Avg. & \cellcolor{orange!10}\textbf{GCT.} & \cellcolor{orange!10}\textbf{MPR.} & \cellcolor{orange!10}\textbf{LI.} & \cellcolor{yellow!10}\textbf{FA.} & \cellcolor{yellow!10}\textbf{OP.} & \cellcolor{yellow!10}\textbf{TS.} \\ \hline
\rowcolor{navyblue!5} \multicolumn{1}{l|}{\textcolor{black}{\textit{Proprietary Models (API)}}} & & & & & & & &\\

\textsuperscript{\dag}Gemini-3.1-Pro \textsuperscript{\ddag} & 1 &
    \cellcolor{oai-gray-600}{73.8} & \cellcolor{oai-gray-600}{59.1} &
    \cellcolor{oai-gray-600}{83.4} & \cellcolor{oai-gray-600}{86.8} & 73.2
  &
    67.9 & \cellcolor{oai-gray-600}{72.1} \\

  \textsuperscript{\dag}GPT-5.4-High \textsuperscript{\ddag} & 2 & 72.1 &
    58.4 & 82.8 & 81.1 & \cellcolor{oai-gray-600}{76.2} & 65.5 & 68.8 \\

  \textsuperscript{\dag}Gemini-3.1-Flash-Lite \textsuperscript{\ddag} & 3 &
   69.4 & 55.0 & 78.3 & 81.3 & 58.3 & \cellcolor{oai-gray-600}{77.7} & 66.3 \\

  \textsuperscript{\dag}GPT-5 \textsuperscript{\ddag} & 4 & 69.4 & 58.4 &
    83.0 & 81.5 & 75.3 & 60.2 & 58.1 \\

  \textsuperscript{\dag}GPT-5.4 \textsuperscript{\ddag} & 5 & 67.3 & 54.5 &
    79.6 & 83.0 & 66.4 & 63.4 & 57.0 \\

  \textsuperscript{\dag}Gemini-2.5 Pro \textsuperscript{\ddag} & 6 & 67.1 &
    54.4 & 80.7 & 83.8 & 65.5 & 60.2 & 58.1 \\

  \textsuperscript{\dag}o4-mini \textsuperscript{\ddag} & 7 & 66.8 & 51.0 &
    73.7 & 82.4 & 68.5 & 65.0 & 60.4 \\

  \textsuperscript{\dag}Qwen3-VL-Plus & 8 & 58.1 & 51.3 & 64.3 & 73.6 &
    61.3 & 50.4 & 47.7 \\

  \textsuperscript{\dag}Gemini-2.5 Flash \textsuperscript{\ddag} & 9 & 55.3
    & 41.5 & 66.8 & 73.3 & 52.1 & 50.4 & 47.7 \\ \hline

\rowcolor{navyblue!5} \multicolumn{1}{l|}{\textcolor{black}{\textit{Open-source Instruct Models}}} & & & & & & & &\\
\textsuperscript{\dag}Qwen3-VL-235B-A22B-Instruct &3 &60.7 &52.3 &66.6 &\cellcolor{oai-gray-300}{78.8} &55.5 &53.0 &58.1\\
\textsuperscript{\dag}Qwen3-VL-32B-Instruct &4&59.0 &50.0 &64.3 &76.7 &55.5 &53.1 &54.6\\
\textsuperscript{\dag}Qwen3-VL-30B-A3B-Instruct & 9&52.7 &42.1 &57.6 &75.5 &46.2 & 49.6& 45.3\\
\textsuperscript{\dag}Qwen3-VL-8B-Instruct & 8&53.3 &41.5 &56.3 &73.1 &45.4 &54.9 &48.8\\
InternVL3.5-30B-A3B &5&55.9 & 48.5&59.5 &74.0 &50.0 & 51.3&52.3\\
InternVL3.5-14B &6&55.3 & 44.6&63.6 &72.1 &44.5 &52.2 &54.7\\
InternVL3.5-8B &7&53.9 &44.4 &57.9 &69.0 &46.6 &53.1 &52.3\\
LLaVA-OneVision-7B &11&50.4 & 40.5 & 57.3 & 60.2 & 44.5 & 49.9 & 50.3\\
LLaVA-OneVision-72B &2&61.3&52.8 &64.2 &68.6 &\cellcolor{oai-gray-300}{60.1} &\cellcolor{oai-gray-300}{58.4} &\cellcolor{oai-gray-300}{61.0}\\
LLaVA-Video-7B &10& 50.9 &55.4 &61.1 &67.9 &49.2 &38.9 &32.6\\
LLaVA-Video-72B &1& 64.9& \cellcolor{oai-gray-300}{57.9} &\cellcolor{oai-gray-300}{70.3} &75.2 &56.7 &58.4 &50.9\\ \hline

\rowcolor{navyblue!5} \multicolumn{1}{l|}{\textcolor{black}{\textit{Open-source Reasoning Models }}} & & & & & & & &\\
\textsuperscript{\dag}Qwen3-VL-235B-A22B-Thinking & 1& 57.9&53.8 &62.4 &74.0 &60.9 &51.3 &45.3\\
\textsuperscript{\dag}Qwen3-VL-32B-Thinking &2 & 55.9&49.5 &64.0 &75.7 &59.7 & 42.5&44.2\\
\textsuperscript{\dag}Qwen3-VL-30B-A3B-Thinking & 3&52.1 &41.5 &59.9 &75.0 &46.6 &39.8 &50.0\\
\textsuperscript{\dag}Qwen3-VL-8B-Thinking &4& 51.4 & 42.6 & 58.3 &70.7 &48.3 &40.7 &47.7\\
\textsuperscript{\dag}GLM-4.5V-Thinking &5& 45.1 &28.7 &53.5 &65.5 &41.2 &42.5 & 39.5 \\
\end{tabular}}

    \end{minipage}
    \hfill
    \begin{minipage}{0.34\textwidth}
        \centering
        \includegraphics[width=\linewidth]{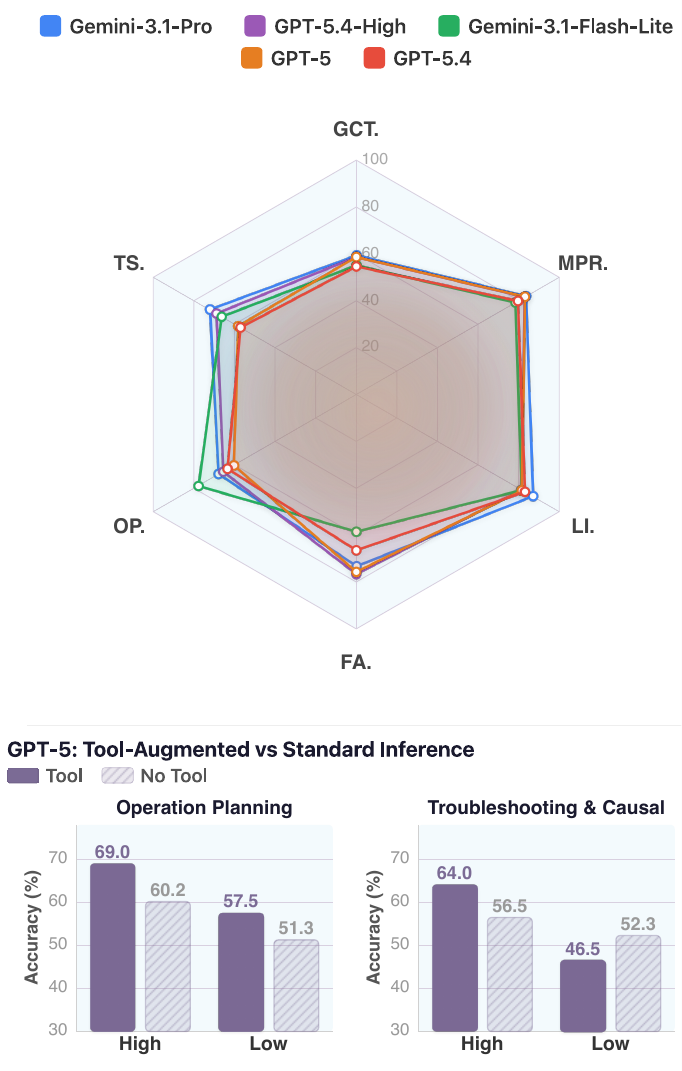}
    \end{minipage}
    
   \caption{\textbf{Evaluation on SFI-Bench.} The task names have been abbreviated for improved readability. \textbf{Left:} Avg. is macro-average accuracy. \colorbox{oai-gray-600}{Dark gray} indicates the best result among all models and \colorbox{oai-gray-300}{light gray} indicates the best result among open-source models. \colorbox{orange!10}{Orange} highlights spatial reasoning tasks, while \colorbox{yellow!10}{yellow} highlights functional reasoning tasks. \textsuperscript{\dag} indicates calling model with official API. \textsuperscript{\ddag} indicates model with internet search tool for last two tasks.  \textbf{Top Right:} Radar plot of the best-performing models. \textbf{Bottom Right:} GPT-5 performance on the last two tasks under different reasoning modes and with/without web search tool.}

    \label{tab:eval}
\vspace{-3mm}
    
\end{figure*}

\subsection{Main Results}

Table \ref{tab:eval} presents the overall results on SFI-Bench.

\vspace{0.05cm}
\noindent\textbf{Proprietary Models.}  
Among proprietary systems, reasoning-enabled variants consistently yield substantial gains, indicating that the improvements are primarily driven by enhanced reasoning capabilities. Gemini-3.1-Pro achieves the strongest overall performance, while GPT-5.4-High outperforms GPT-5.4, and Gemini-2.5-Pro similarly surpasses Gemini-2.5-Flash.
 Across all models, global conditional counting emerges as a key bottleneck, revealing persistent limitations in compositional and logical reasoning. While leading proprietary models exhibit strong capabilities in spatial cognitive map construction, their performance on functional reasoning tasks remains comparatively weaker. This gap becomes more pronounced on the two knowledge-grounded tasks (\cref{tab:eval} bottom right): GPT-5 equipped with web search tool significantly outperforms its offline counterpart under high reasoning budgets.  

However, web search tool use introduces additional noise and can degrade performance when reasoning capacity is limited. Even within GPT-5, the low-reasoning variant performs worse with web search than without it on troubleshooting tasks. A similar trend is observed for Gemini-2.5-Flash and Qwen3-VL-Plus, where reasoning-enabled variants underperform their instruction-tuned counterparts. These findings suggest that strong reasoning ability is a prerequisite for effective tool use.

\vspace{0.05cm}
\noindent\textbf{Open-source Models.}  
Among open-source systems, video-based models such as LLaVA-Video-72B achieve strong spatial reasoning performance, even surpassing Gemini-2.5-Flash on several tasks. 
Nevertheless, the overall open-source ecosystem remains substantially behind proprietary models. 
Global conditional counting persists as the primary bottleneck, while layout inference is comparatively easier. 
Models without internet access must rely solely on parametric knowledge, yielding accuracies near 50\% on functional reasoning tasks.

Notably, open-source reasoning models show minimal improvement over instruct counterparts.

\finding{1}{Spatial and layout understanding can be improved with effective reasoning. However, additional information (e.g., external tools) may introduce noise that degrades performance for weaker reasoning models. State-of-the-art open-source multimodal models trained with RLVR fail to effectively transfer reasoning capabilities from mathematical domains to spatial–functional tasks.}

\begin{figure*}[!htb]
    \centering
    \includegraphics[width=\linewidth]{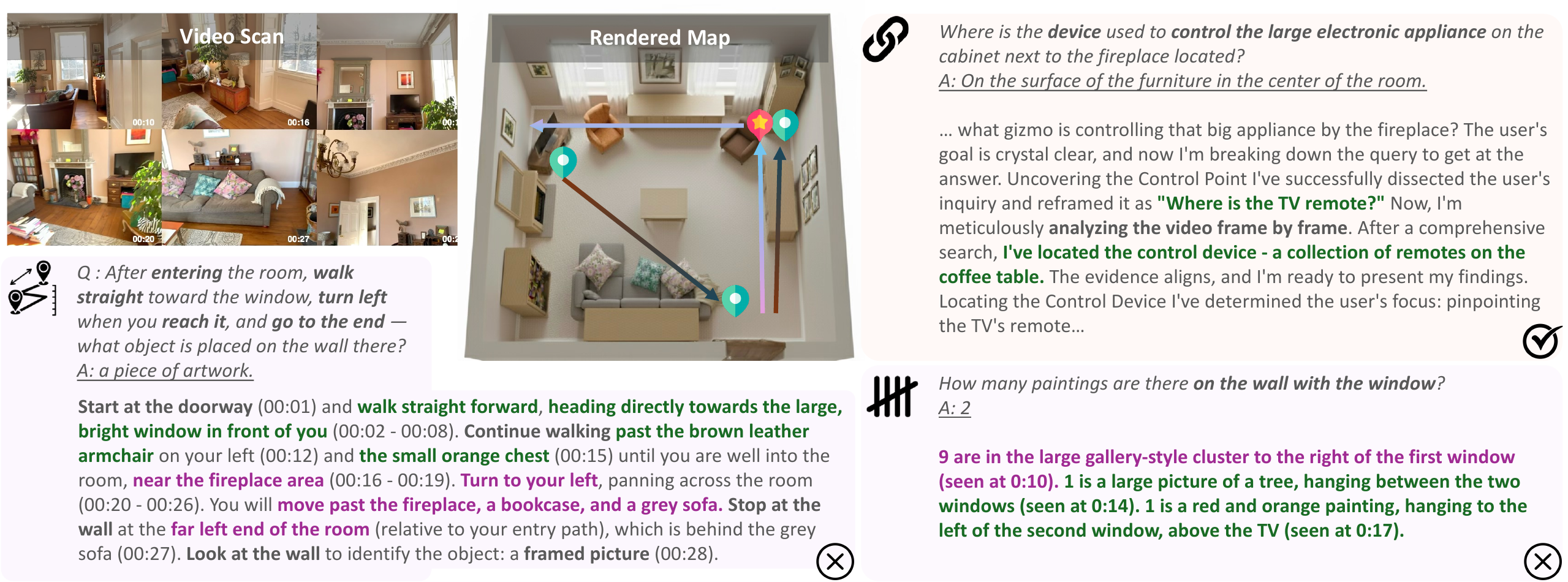}
  \caption{\textbf{Human-annotated analysis of MLLM reasoning chain.} 
The central panel visualizes a conceptual cognitive map reconstructed from the video. Arrows denote navigation trajectories required to answer the question: red arrows indicate the ground-truth path in the environment, while blue arrows represent the model’s inferred reasoning path. 
Experts annotate errors by projecting the model’s reasoning steps onto the spatial layout, revealing discontinuities (e.g., abrupt jumps between locations) and hallucinated transitions. Key steps are labeled, with \textcolor{teal!80!black}{\textbf{correct}} and \textcolor{mymagenta}{\textbf{incorrect}} reasoning highlighted, exposing both successful spatial reconstruction and common failure modes.}
    \label{fig:case_study}
\vspace{-3mm}
\end{figure*}

\section{Limitation of Current MLLMs}

We investigate how modern MLLMs reason about \textit{space} and \textit{functionality}—specifically, how they construct and utilize cognitive spatial maps and functionality maps when solving complex multimodal reasoning tasks. 
By examining both reasoning traces and systematic failure patterns, we aim to uncover the mechanisms and limitations underlying spatial understanding, functional inference, and their interaction in current vision–language systems.

\begin{figure*}[!htb]
    \centering
    \includegraphics[width=\linewidth]{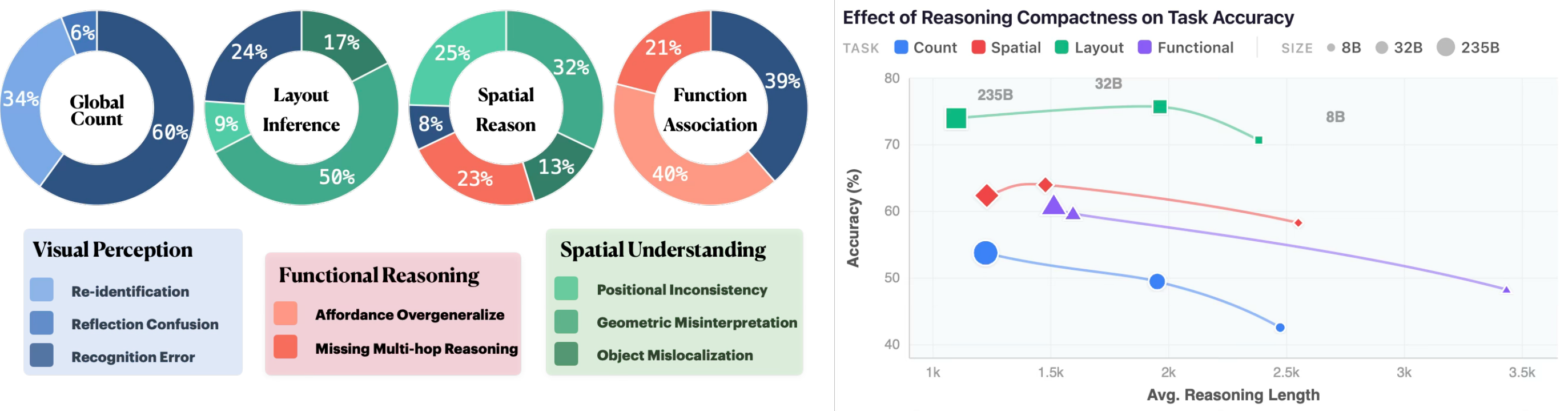}
    \caption{\textbf{Left:} Human analysis categorizing task-specific failures. \textbf{Right:} Relationship between reasoning compactness and task accuracy. Colors denote task types, while point size encodes Qwen3-VL model scale (8B, 32B, 235B). Regression lines indicate that larger models tend to produce shorter reasoning chains, which consistently correlate with higher accuracy.}
    \label{fig:error_reasonacc}
    \vspace{-4mm}
\end{figure*}

\subsection{Probing via Reasoning Traces}

To probe how state-of-the-art MLLMs internalize spatial and functional structure, we analyze the reasoning traces of Gemini-2.5 Pro.
Modern MLLMs often employ reinforcement-learning-enhanced reasoning to produce explicit textual chains of thought~\cite{comanici2025gemini,gpt4o,guo2025deepseek}, which provide a valuable lens into their decision processes~\cite{wang2025multimodal,marjanovic2025deepseek}. 

\vspace{0.05cm}
\noindent\textbf{Case Study.} \cref{fig:case_study} presents representative examples of spatial and functional reasoning chain. In successful cases, the model decomposes questions into manageable subtasks (e.g., locating relevant regions, aligning cross-view cues, inferring relationships), integrates multi-frame evidence, and constructs a coherent internal scene representation. 
These behaviors suggest early-stage cognitive spatial mapping, reflecting an emergent understanding of object permanence and spatial continuity. 

However, performance degrades on multi-hop reasoning.
In cross-view reasoning (bottom left), the model exhibits \emph{fragmented spatial continuity}, "jumping" between distant objects (e.g., from a chest to a fireplace) without maintaining consistent self-location. 
In conditional counting (bottom right), the model \emph{misinterprets contextual modifiers} (e.g., “on the wall with the window”), revealing a gap between linguistic conditioning and geometric grounding.

General functional reasoning tasks amplify these challenges. 
While the model can recognize object affordances (e.g., linking a coffee machine with mugs), it struggles with multi-step reasoning, such as identifying prerequisites (“fill with water”) or understanding control dependencies (“the remote operates the TV”). 
In causal troubleshooting, it fails to integrate visual cues with external knowledge, often generating superficial or hallucinated explanations instead of grounded diagnoses.

\vspace{0.05cm}
\noindent\textbf{Failure Mode Analysis.} \label{sec:failuremode} To identify the sources of failure in state-of-the-art MLLMs' reasoning on \textit{spatial–functional intelligence}, we analyze 120 erroneous samples produced by Gemini-2.5-Pro across the first four tasks (30 samples per task). We identify three main categories of failure modes (see Appendix for details):

\begin{enumerate}
    \item \textcolor{blue!90!black}{\textbf{Visual Perception:}}
    Failures in object recognition and visual interpretation, including \textit{missing objects}, \textit{misclassification}, \textit{attribute mislabeling}, \textit{re-identification failure}, and \textit{reflection confusion}.

    \item \textcolor{green!60!black}{\textbf{Spatial Understanding:}}
    Errors in maintaining spatial consistency and inferring spatial relation, such as \textit{positional inconsistency}, \textit{geometric misinterpretation}, and \textit{object mislocalization}.

    \item \textcolor{red!70}{\textbf{Functional Reasoning:}}
    Failures related to the model’s ability to understand functional relationships and perform grounded, compositional reasoning.
    This includes: \textit{Affordance overgeneralization}, where the model assumes functional relationships based on commonsense (e.g., assuming any remote controls a TV) without verifying the specific context; and \textit{Missing multi-hop reasoning}, where the model fails to complete complex multi-step inferences over functional chain of objects.
\end{enumerate}

\cref{fig:error_reasonacc} (left) shows that error patterns across tasks. It highlights that visual perception issues are common across all tasks, pointing to the ongoing difficulty of video understanding for existing MLLMs. Spatial understanding is critical for spatial and layout reasoning tasks, while functional reasoning errors primarily affect spatial reasoning and functional association, where the model often overlooks fine-grained functional relationships and struggles with object recognition. This trend reveals a clear progression: 
\finding{2}{As tasks become more cognition-intensive, errors shift from perception to reasoning and spatial–functional grounding. Models fail to maintain consistent spatial localization, exhibiting discontinuous spatial memory and hallucinated transitions.}

\begin{figure}[t]
    \centering
    \includegraphics[width=\linewidth]{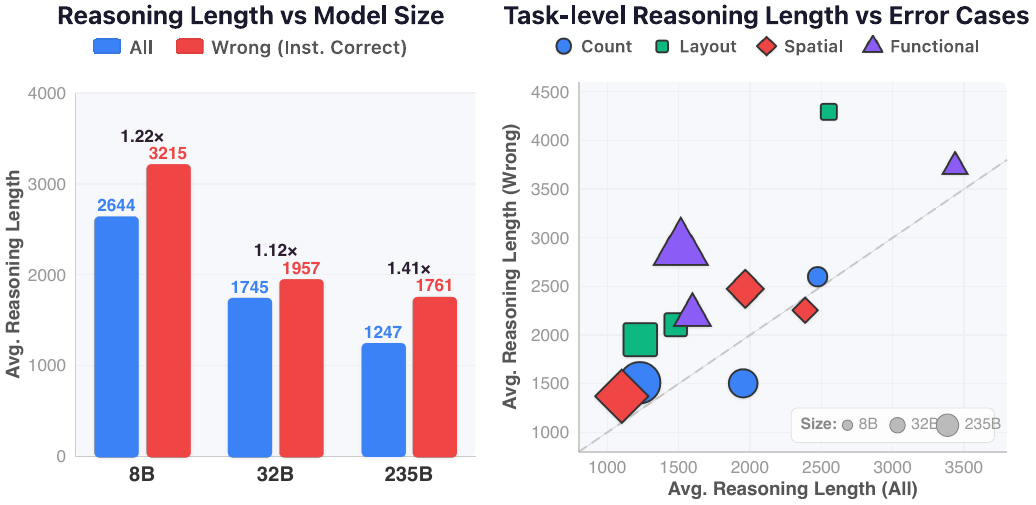}
    \caption{
    \textbf{Left:} Average reasoning lengths across Qwen3-VL model scales (8B, 32B, 235B). \textit{Wrong} denotes cases where the reasoning model failed while the instruction model succeeded. 
    \textbf{Right:} Task-level reasoning length comparison across models. The Y-axis shows the average token length for \textit{Wrong} cases, and the X-axis for all cases. 
    Points above the diagonal and that lie farther away indicate tasks where models tend to produce unnecessarily long and erroneous reasoning chains.}
    \label{fig:reasonleng}
    \vspace{-4mm}
\end{figure}

\subsection{Limited Reasoning Gains from RLVR}

\cref{tab:eval} reveals that open-source reasoning variants trained with reinforcement learning with verifiable rewards (RLVR) offer only marginal improvements over their instruction-tuned counterparts. 
To investigate this, we analyze the reasoning lengths of the Qwen3-VL family across model scales and tasks (Fig.~\ref{fig:error_reasonacc} right; \cref{fig:reasonleng}). 
Across all settings, longer reasoning does not correlate with higher accuracy. 
Notably, samples that were answered correctly in the non-reasoning mode but failed in the reasoning mode exhibit substantially longer reasoning chains—1.41×, 1.12×, and 1.22× longer than the global averages for the 235B, 32B, and 8B models, respectively (Fig.~\ref{fig:reasonleng} left). 
The above results suggests that excessive reasoning often leads to over-explanation and semantic drift rather than deeper inference.

As shown in \cref{fig:error_reasonacc} (right), larger models (235B) generate shorter yet more effective reasoning, achieving higher accuracy through compact, well-grounded planning. 
Smaller models (8B), by contrast, rely on verbose but shallow reasoning to compensate for weaker internal representations. 
Overall, reasoning length proves to be an unreliable indicator of reasoning quality—beyond a moderate range ($\approx$1.2–1.5k tokens), linguistic noise dominates.

At the task level (see \cref{fig:reasonleng} and \cref{fig:error_reasonacc}), \textit{functional association} shows the most pronounced overextension effect, where longer reasoning chains lead to increased failure rates. This indicates that the model tends to overthink, resulting in overly long reasoning chains that introduce noise. \textit{Spatial reasoning} remains relatively stable across lengths, while \textit{layout inference} shows occasional inflation caused by repetitive scene descriptions. 
\textit{Counting} is the most concise and stable task, reflecting strong visual grounding.

\finding{3}{Effective reasoning is not about producing longer explanations but about maintaining alignment with perceptual evidence. 
Concise, grounded, and semantically coherent reasoning yields the most reliable improvements across tasks and model scales.}

\section{Determinants of Performance on SFI-Bench}

\subsection{Temporal Consistency Dependence}

\begin{table}[t]
\centering

\resizebox{0.8\columnwidth}{!}{%
\begin{tabular}{lccccc}
\textbf{SR} & \textbf{Overall} & \textbf{Count} & \textbf{Layout} & \textbf{Spatial} & \textbf{Func.} \\
\toprule
0 & \textbf{75.5} & 60.0 & \textbf{88.0} & \textbf{82.0} & 72.0 \\
25\% & \textbf{75.5} & \textbf{62.0} & \textbf{88.0} & 80.0 & 72.0 \\
50\% & 71.5 & 50.0 & 84.0 & 80.0 & 72.0 \\
75\% & 68.5 & 50.0 & 78.0 & 76.0 & 70.0 \\
100\% & 75.0 & 58.0 & 86.0 & 78.0 & \textbf{78.0} \\
\end{tabular}%

}
\caption{\textbf{Effect of frame order shuffling rate (SR) on task accuracy (\%)}
Higher SR indicates stronger temporal disruption.}
\label{tab:temporal_shuffle}
 \vspace{-5mm}
\end{table}
Humans construct cognitive maps by integrating perceptual inputs over continuous time.
Neuroscience studies show that temporal continuity is essential for spatial learning, as hippocampal and entorhinal systems encode spatial relations through sequential experience \cite{o1978hippocampus, epstein2017cognitive}.
To test whether large multimodal models share this property, we progressively shuffled the input frame order for GPT-5 at different shuffle ratios (SR).
As shown in Table~\ref{tab:temporal_shuffle}, performance remained largely stable, with only moderate degradation in counting tasks.
This experiment was conducted on a subset of 200 samples (50 per task type), and the results indicate that MLLMs rely primarily on aggregated visual evidence rather than temporally coherent dynamics, forming a static spatial abstraction rather than a time-dependent cognitive map.

\subsection{Visual Grounding vs. Textual Descriptions}
\begin{table}[ht]
\centering

\resizebox{0.65\columnwidth}{!}{%
\begin{tabular}{lcccc}
\textbf{Input} & \textbf{Count} & \textbf{Layout} & \textbf{Spatial} & \textbf{Func.} \\
\toprule
Visual & 58.4 & 83.0 & 81.5 & 75.3 \\
Caption-only & 57.2 &51.6 &55.4& 67.6 \\
\end{tabular}%

}
\caption{\textbf{Visual vs. Caption-Based Input.} GPT-5 performance under full video input and caption-based structured descriptions.}

\label{tab:visualdepend}
\vspace{-3mm}
\end{table}

To assess whether cognitive map construction relies on global visual signals or can be inferred from structured textual descriptions, we evaluate GPT-5 on 200 samples under two conditions: (1) full visual input, and (2) structured textual input derived from captions, similar to socratic models approach \cite{zeng2022socratic}. The captions are generated by the same model by first processing the video and producing detailed descriptions of object attributes, relations, and layout.

As shown in \cref{tab:visualdepend}, performance remains strong with visual input but drops consistently when only structured textual descriptions are provided. The degradation is most pronounced in tasks requiring spatial and layout understanding.  

These results suggest that even detailed, model-generated textual descriptions are insufficient for accurate cognitive map construction, highlighting the critical role of direct visual grounding.

\section{Related Work}

\paragraph{Multimodal Large Language Models.} Multimodal large language models (MLLMs) have achieved significant success~\cite{hurst2024gpto,liu2023visual,li2024llava,bai2023qwen,tong2024cambrian,team2023gemini,chen2024internvl,wang2024qwen2vl,li2023blip2, xin2025lumina} by effectively integrating rich visual semantics from vision encoders~\cite{radford2021learning,zhai2023sigmoid,tschannen2025siglip} with the sophisticated reasoning abilities of LLMs~\cite{brown2020language,touvron2023llama,yang2025qwen3}.
As a natural extension of this momentum, this architecture has been adapted for the temporal domain, leading to the rapid development of video-based MLLMs~\cite{li2024llama,li2024llava,zhang2024video,song2024moviechat,bai2025qwen2,zhu2025internvl3,zhang2023video,li2023videochat,zohar2024apollo,marafioti2025smolvlm} sampling separate frames in video and concatenating their per frame features, which are expected to serve as foundation for real-world embodied agents \cite{kim2024openvla, yang2024virl, chen2026roborouter, yang2026evotool, ni2025swiftvla, team2025gigabrain, team2025gigaworld, wang2025embodiedreamer, dong2025emma}. 
Concurrently, evaluation is also fundamental to drive model evolution.
Massive VQA benchmarks have emerged~\cite{awal2024vismin,yue2024mmmu,tong2024eyes,tong2024cambrian,fu2025video,mangalam2023egoschema,majumdar2024openeqa, yang-etal-2025-magic, yang2025multimodal}, to test knowledge recall and semantic understanding of MLLMs. However, most existing benchmarks primarily focus on content or activity level understanding, ignoring a fundamental primitive: spatial layout within the video.
Recent works are beginning to draw the community's attention to this gap, emphasizing the need to examine the spatial reasoning and underlying world understanding capabilities of MLLMs in both images and videos~\cite{yang2024think, yin2025spatial, yeh2025seeing}.

\paragraph{Visual Spatial Intelligence.} Spatial intelligence refers to the ability to perceive, represent, and reason about spatial relationships, a foundational concept in cognitive psychology~\cite{shepard1986mental, gardner2011frames, newcombe2004spatial}. Humans excel at building mental maps of their surroundings, performing egocentric--allocentric transformations, and leveraging spatial memory for navigation and problem-solving ~\cite{yang2024think, yang2025cambrian}. For MLLMs to perceive and interact with the physical world, a robust understanding of spatial relationships from visual inputs is crucial. Recognizing this requirements, recent research has advanced in two parallel directions: designing physically and spatially grounded benchmarks ~\cite{yang2024think,ramakrishnan2024does,yin2025spatial,majumdar2024openeqa,yeh2025seeing,li2025sti,xu2025multi,team2025gemini,brown2025shortcuts}, and developing new methods to enhance the spatial reasoning capabilities of MLLMs~\cite{yang2025mindjourney,ma2025spatialreasoner,ouyang2025spacer,du2024embspatial,chen2024spatialvlm,cheng2024spatialrgpt,cai2024spatialbot,liu2024coarsecorrespondenceelicit3d,li2024topviewrs,zhu2024llava3d,ray2025sat,fan2025vlm,yang2025visualspatialtuning}.

While prior efforts focused on spatial reasoning in MLLMs---often limited to understanding spatial layouts from video---our work extends this scope by incorporating functional reasoning over objects and their potential interactions. Moreover, we evaluate the role of web search, aligning with recent video-based deep research~\cite{liu2026watching} that integrates external knowledge for comprehensive reasoning.

\section{Conclusion and Future Work}

We introduce SFI-Bench, a benchmark that moves multimodal evaluation beyond basic perception to spatial reasoning and functional understanding. While current MLLMs perform well on perception, they struggle with spatial memory, functional knowledge integration, and linking perception to external knowledge. Open-source models also show limited transfer of reasoning from visual math tasks to spatial–functional settings. We find that web search is critical, with online models outperforming offline variants. These results highlight the need for stronger spatial memory, more robust compositional reasoning, and better integration of external knowledge for building intelligent spatial agents.
\clearpage

{
    \small
    \bibliographystyle{ieeenat_fullname}
    \bibliography{main}
}

\appendix
\setcounter{page}{0}
\setcounter{section}{0}
\maketitlesupplementary

\section{Data Curation Process Details}
\label{app:app_datacurationdetails}

This section provides additional details on the dataset construction process, including the metadata format, prompting setup, and human annotation pipeline. Unless otherwise specified, Gemini-2.5-Pro is used as the default MLLM assistant. For clarity, the prompts shown here are lightly streamlined.

\paragraph{Metadata Generation.}
We begin by generating structured metadata for each video using an MLLM. The metadata includes global scene descriptions, object instances with timestamps, category labels, fine-grained attributes (\eg, color, material, brand), spatial relations, and functional tags. An example (simplified for readability) is shown in \cref{fig:json_output_full}. The system prompt used for metadata generation is provided in \cref{fig:meta_info_system_prompt}. Outputs from multiple passes are merged and subsequently verified against the original video.

\paragraph{Task Templates.}
Given high-quality metadata, task-specific questions are produced using carefully designed templates. The templates for each task are shown in: \cref{fig:globalcountingtemplate} (global conditional counting), \cref{fig:spatialreasoningtemplate} (cross-view multi-hop path reasoning), \cref{fig:layouttemplate} (layout inference), and \cref{fig:functionalasscociationtemplate} (functional association). Each template is tailored to ensure consistency, task fidelity, and sufficient reasoning complexity.

\begin{figure*}[t]
    \centering
    \includegraphics[width=\linewidth]{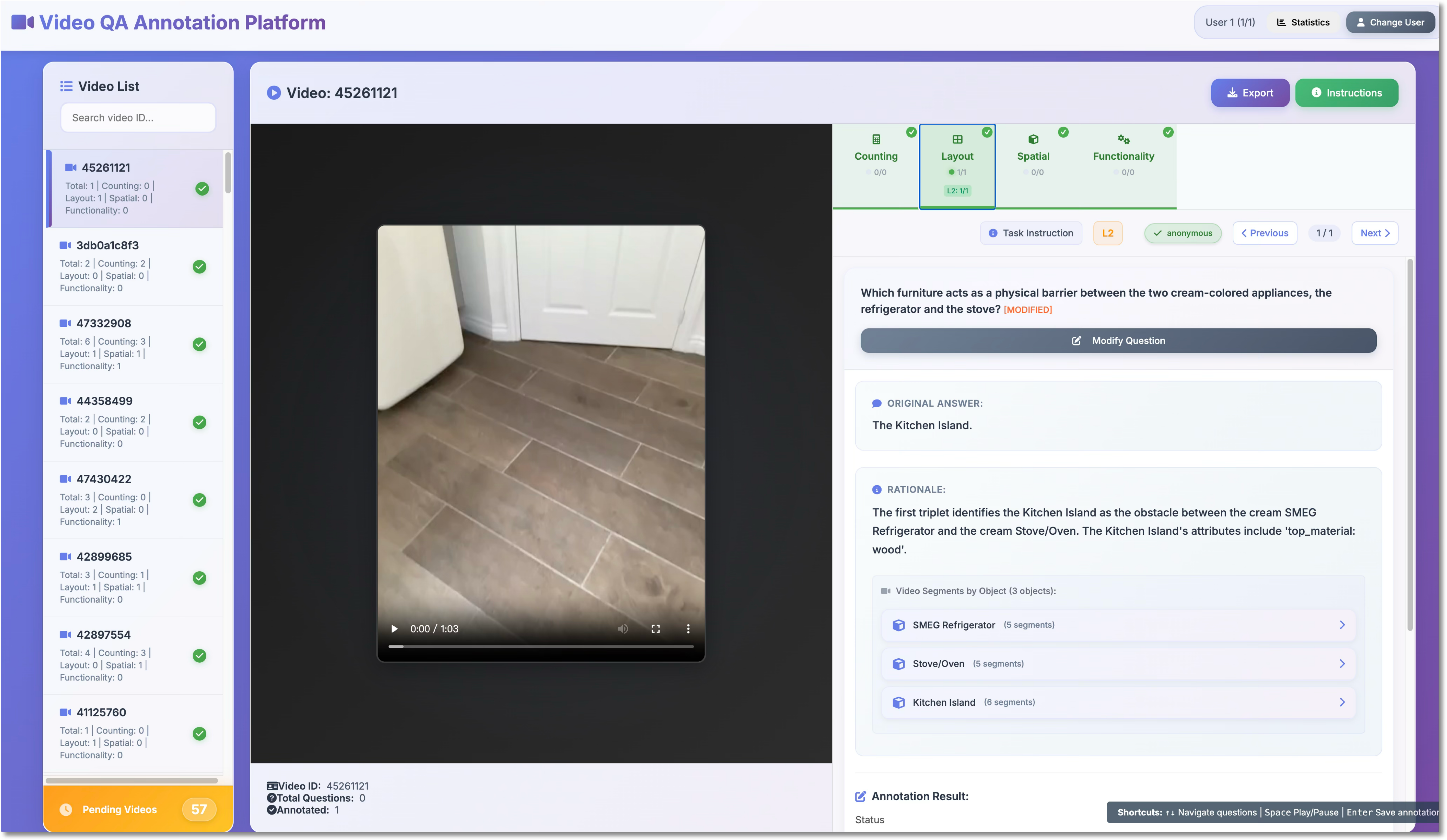}
    \caption{Annotation platform for human verification of questions and answers.}
    \label{fig:platform}
\end{figure*}

\paragraph{Human Verification.}
The dataset is curated by 11 experienced annotators, all trained machine learning engineers. Each question is manually reviewed by at least one expert to ensure correctness and alignment with the associated video. Annotations are performed using an HTML-based platform that displays the question-generation reasoning chain, relevant objects with timestamps, and the corresponding video interface, as illustrated in \cref{fig:platform}.

\section{Evaluation Setup}

All videos are preprocessed to a fixed resolution of \textbf{720p} and \textbf{24\,fps} before being uploaded to any model. This standardization reduces data size and ensures consistent visual quality across systems, preventing performance differences arising from input variability.

For proprietary models, we evaluate the Qwen3-VL family through the official \textbf{Aliyun API}, Gemini models via the \textbf{Google Cloud API}, and GPT models using the \textbf{OpenAI API}. All API-based experiments were conducted during the first two weeks of \textbf{September 2025} to ensure fair comparison under stable model deployments.

For open-source models, we adopt the \textbf{VLMEvalKit} framework to provide a unified inference pipeline and standardized evaluation protocol. This setup enables consistent multi-image and video processing across LLaVA-based, InternVL-based, and GLM-based models.

To promote reproducibility, we will release our \textbf{evaluation scripts}, \textbf{preprocessing pipeline}, and the full \textbf{SFI-Bench dataset} upon publication.

\section{Influence of Reasoning Budget}

\begin{figure*}[t]
    \centering
    \includegraphics[width=\linewidth]{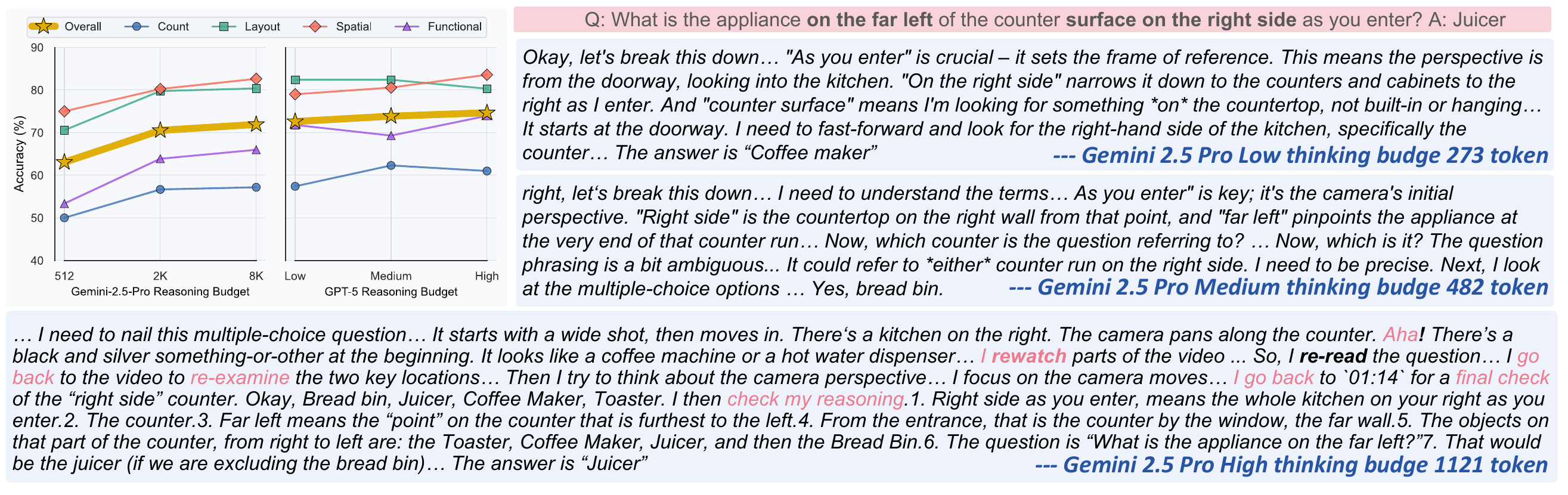}
    \caption{\textbf{(Left)} Performance of Gemini-2.5 Pro and GPT-5 under different reasoning budgets. \textbf{(Right)} Illustration of how increased reasoning budgets alter reasoning patterns: higher budgets reveal frequent behaviors such as \textcolor{red!40!black}{self-rechecking} and complex reasoning chains.}
    \label{fig:reason_budget}
    \vspace{-4mm}
\end{figure*}

In the main paper, we analyzed open-source model performance based on model scale without explicitly controlling the reasoning parameter. To investigate whether RL-based optimization improves reasoning on SFI-Bench, we now \textbf{actively control the \textit{reasoning budget}}---the maximum token allowance for the model's reasoning chain---on proprietary models. By varying the budget (see \cref{fig:reason_budget}), we can directly assess how changes in reasoning depth affect performance across all SFI-Bench tasks.

Both GPT-5 and Gemini-2.5-Pro show accuracy gains with increased reasoning budgets, indicating that extended reasoning allows for better integration of spatial and functional cues. However, the performance curve flattens beyond approximately 2k tokens, as Gemini-2.5-Pro does not utilize reasoning beyond this point, reaching its limit in reasoning depth.

By examining the reasoning content, we observe that longer reasoning budgets tend to reveal interesting patterns, such as frequent self-checking and the development of complex, ordered plans for task-solving. Overall, the results suggest that reasoning efficiency, rather than token capacity alone, drives performance. A moderate reasoning budget (around 2k tokens) offers an optimal balance between expressivity and stability. Beyond this point, further expansion yields diminishing returns, as effective models reach a point of sufficient problem-solving without continuing to reason indefinitely.

\section{Detailed Failure Mode Categorization}

Below is the detailed failure mode categorization for the analysis introduced in \cref{sec:failuremode}:

\begin{enumerate}
    \item \textcolor{blue!90!black}{\textbf{Visual Perception:}}
    Failures related to object recognition and visual data interpretation. This includes:
    \textit{Missing objects}, where the model overlooks visible entities;
    \textit{Object misclassification}, where objects are assigned the wrong labels;
    \textit{Attribute mislabeling}, where attributes such as color, size, or brand are incorrectly identified;
    \textit{Re-identification failure}, where the model mistakenly counts the same object multiple times when viewed from different perspectives; and
    \textit{Reflection confusion}, where the model misinterprets mirror reflections as real objects.

    \item \textcolor{green!60!black}{\textbf{Spatial Understanding:}}
    Errors related to the model's ability to maintain consistent and accurate spatial representations. This includes:
    \textit{Positional inconsistency}, where objects shift positions or lose continuity across frames (\eg, teleportation effects);
    \textit{Geometric misinterpretation}, where the model fails to infer correct geometric relationships between objects (\eg, alignment or linearity); and
    \textit{Object mislocalization}, where the model places objects in incorrect locations, such as confusing left/right or near/far positioning.

    \item \textcolor{red!70}{\textbf{Functional Reasoning:}}
    Failures related to the model's ability to understand functional relationships and perform grounded, compositional reasoning. This includes:
    \textit{Affordance overgeneralization}, where the model assumes functional relationships based on commonsense (\eg, assuming any remote controls a TV) without verifying the specific context; and
    \textit{Missing multi-hop reasoning}, where the model fails to complete complex multi-step inferences over functional chains of objects.
\end{enumerate}


\begin{figure*}[t]
\centering
\begin{lstlisting}[language=promptJson]
{
  "video_file": "41069048.mp4",
  "scene_overview": {
    "room": "Bathroom",
    "style": "Modern, clean, functional, hotel-like",
    "palette": "White, grey-blue, chrome"
  },
  "objects": [
    {
      "id": "toilet_001",
      "class": "Toilet",
      "prominent_ts": "00:10",
      "vis_segments": [ {"start": "00:00", "end": "00:02"}, {"start": "00:08", "end": "00:13"} ],
      "attributes": { "type": "Two-piece", "mat": "ceramic", "color": "white" },
      "state": { "lid": "closed", "condition": "clean" },
      "location": "Positioned between trash can and bathtub, against wall",
      "functionality": { "primary": "Waste disposal", "secondary": [] },
      "relations": {
         "type": "functional_group",
         "related": ["toilet_brush_001", "trash_can_001", "toilet_paper_holder_001"]
      }
    },
    {
      "id": "bathtub_001",
      "class": "Bathtub",
      "prominent_ts": "00:19",
      "vis_segments": [ {"start": "00:11", "end": "00:20"} ],
      "attributes": { "type": "Shower-tub combo", "mat": "acrylic", "color": "white" },
      "state": { "fill_level": "empty", "condition": "clean" },
      "location": "Adjacent to sink and toilet, against wall",
      "functionality": { "primary": "Bathing/Showering", "secondary": [] },
      "relations": {
         "type": "integrated_system",
         "core_components": ["shower_system_001", "shower_screen_001", "grab_bar_001"]
      }
    }
    ...
  ],
  "spatialLayout": {
    "mainPathway": "Camera moves in circular path: toilet -> wall -> ceiling -> shower/tub -> sink -> door.",
    "relativePositions": "Toilet and towel rail on one side; bathtub and sink on the opposite side.",
    "anomaliesOrAbsences": "A DVD case on the bathroom floor (unusual). Toilet paper holder is empty."
  },
  "functionalEcosystem": {
    "hygiene_and_grooming": {
      "core_objects": ["sink_001", "toilet_001", "bathtub_001", "shower_system_001"],
      "supporting_objects": ["faucet_001", "soap_bar_001", "mirror_001", "towel_radiator_001",
                             "towel_001", "towel_002", "towel_003", "toilet_brush_001", "trash_can_001"],
      "description": "A complete system for personal hygiene, with all necessary fixtures present."
    }
  }
}
\end{lstlisting}
\vspace{-0.2cm}
\caption{Simplified example of the structured metadata generated for each video.}
\label{fig:json_output_full}
\end{figure*}


\begin{figure*}[t]
\centering
\begin{promptbox}{System Prompt: Meta Information Generation}
\vspace{0.2cm}
\textbf{TASK:} Analyze this video and generate a comprehensive, structured JSON representation of the scene and its contents. The goal is to create a rich, machine-readable format suitable for detailed Q\&A and object-level analysis.

\textbf{OUTPUT FORMAT:} \texttt{Pure JSON object only}

\textbf{KEY PRINCIPLES:}
\begin{itemize}[leftmargin=*, nosep]
    \item \textbf{Unique Instance Tracking:} Every discrete object should be a unique entry in the \texttt{objectInventory}. Assign a persistent \texttt{instance\_id} to each object (\eg, \texttt{book\_001}, \texttt{mug\_001}). This ID is crucial for unambiguous reference, even if it looks identical to others or reappears after being hidden.

    \item \textbf{Structured Properties:} Favor structured key-value pairs within an \texttt{attributes} object over a single description string, enabling more precise and queryable information.

    \item \textbf{Temporal and State Awareness:} Accurately document when objects are visible and how their state changes. Use \texttt{visibility\_segments} to track presence over time. Identify the \texttt{most\_prominent\_timestamp}---the moment when the object appears clearest.

    \item \textbf{Functionality and Relationship Analysis:} Analyze and describe each object's primary functionality and its relationships with others. Infer functional connections, especially when objects share brands or have complementary purposes.

    \item \textbf{Accuracy and Completeness:} Be exhaustive. Capture both large furniture and smaller items. If text is legible (\eg, a brand name), capture it. Use \texttt{null} for undetermined values.
\end{itemize}

\vspace{0.2cm}
\textbf{Required JSON structure (with illustrative examples):}
\begin{lstlisting}[language=promptJson]
{
  "sceneOverview": {
    "roomType": "Living Room with Open Kitchen",
    "styleAndAtmosphere": "Modern minimalist, bright with good natural light",
    "mainColorPalette": "Primarily white and natural wood tones, with accents of blue"
  },
  "objectInventory": [
    {
      "instance_id": "phone_charger_001",
      "object_class": "Phone Charger",
      "visibility_segments": [ { "start": "00:08", "end": "01:25" } ],
      "attributes": { "type": "USB-C cable with wall adapter", "color": "white",
                       "brandAndModel": "Apple 20W USB-C Power Adapter" },
      "state": { "connection_status": "plugged into wall outlet" },
      "relational_location": "On the nightstand next to the bed.",
      "functionality_relation": {
        "target_objects": ["iphone_001"], "relationship_type": "power_supply"
      }
    }
  ],
  "spatialLayout": { "mainPathway": "...", "relativePositions": "...", "anomaliesOrAbsences": "..." },
  "functionalEcosystem": {
    "entertainment_zone": {
      "core_objects": ["tv_001", "sofa_001", "remote_control_001"],
      "description": "Integrated entertainment setup"
    }
  }
}
\end{lstlisting}
\end{promptbox}
\vspace{-0.3cm}
\caption{System prompt used to drive the MLLM for meta information generation.}
\label{fig:meta_info_system_prompt}
\end{figure*}


\begin{figure*}[p]
\centering
\begin{promptbox}{Task Template 1: Global Conditional Counting}
\vspace{0.2cm}
\textbf{ROLE:} You are an AI assistant specializing in VQA dataset creation. Your task is to generate a diverse set of natural, high-quality question--answer pairs from structured JSON annotations.

\textbf{OBJECTIVE:} Create questions clearly answerable via exact lookups.

\textbf{INPUT:} Use the provided \texttt{\{\{OBJECT\_INVENTORY\}\}} as the single source of truth.

\vspace{0.15cm}
\textbf{CRITICAL RULES:}
\begin{enumerate}[leftmargin=*, nosep, label=\arabic*.]
    \item \textbf{Exact Attribute Matching:} Questions must be based on full, exact values. No substrings.
    \item \textbf{Answer Value $\ge$ 2:} The integer answer must be 2 or more.
    \item \textbf{Object-Class Specificity:} MUST specify a clear class (\eg, ``chairs'', ``bottles''). Avoid generic ``objects''.
    \item \textbf{Meaningful Filtering:} Conditions must strictly reduce the count (Answer $<$ Total objects of that class).
    \item \textbf{Scene-Specific Focus:} Questions should be grounded in the specific scene, not universal.
    \item \textbf{Natural Phrasing:} Do not expose JSON structure (\eg, use ``wooden'' instead of ``material: wood'').
\end{enumerate}

\vspace{0.15cm}
\textbf{DIFFICULTY LEVELS:}

\textbf{Level 1 -- Single-Condition Counting} \textit{(Count instances matching ONE attribute.)}
\begin{itemize}[leftmargin=*, nosep]
    \item \textbf{Q:} ``How many of the wine bottles are still sealed?''
    \item \textbf{Rationale:} Filters \texttt{wine\_bottle} where \texttt{attributes.state: "sealed"}.
\end{itemize}

\textbf{Level 2 -- Multi-Condition Counting} \textit{(Combine multiple constraints via AND/OR.)}
\begin{itemize}[leftmargin=*, nosep]
    \item \textbf{Q:} ``How many pieces of glassware are both clear and clean?''
    \item \textbf{Rationale:} Filters \texttt{glassware} where \texttt{color: "clear"} AND \texttt{state: "clean"}.
\end{itemize}

\textbf{Level 3 -- Complex Aggregation} \textit{(Grouping, comparison, or set operations.)}
\begin{itemize}[leftmargin=*, nosep]
    \item \textbf{Q:} ``What is the largest number of wine bottles that come from the same brand?''
    \item \textbf{Rationale:} Groups \texttt{wine\_bottle} by \texttt{brandAndModel}, returns the largest group size.
\end{itemize}

\vspace{0.15cm}
\textbf{OUTPUT FORMAT:}
\begin{lstlisting}[language=promptJson]
[
  {
    "question_id": "scene_id_XX_Y_cnt",
    "level": <1, 2, or 3>,
    "question_text": "<Natural question>",
    "question_type": "count",
    "rationale": "<Step-by-step filtering logic>",
    "visibility_segments": [ ... ],
    "generated_by": "llm-creative"
  }
]
\end{lstlisting}
\end{promptbox}
\vspace{-0.2cm}
\caption{System prompt for generating Global Conditional Counting tasks.}
\label{fig:globalcountingtemplate}
\end{figure*}


\begin{figure*}[p]
\centering
\begin{promptbox}{Task Template 2: Cross-View Multi-hop Path Reasoning}
\vspace{0.2cm}
\textbf{ROLE:} You are an AI expert in spatial reasoning. Your task is to generate complex, path-dependent Question-Answer pairs based on video JSON annotations.

\textbf{TARGET TASK:} \textit{Cross-View \& Path-Dependent Localization}. Questions must test a model's ability to construct a 3D mental map and follow multi-step spatial chains without explicit target naming.

\vspace{0.15cm}
\textbf{1.\ CORE INSTRUCTIONS (The ``Path'' Logic):}
\begin{itemize}[leftmargin=*, nosep]
    \item \textbf{Select a Target:} The answer object (hidden from the question text).
    \item \textbf{Identify Landmarks:} Select anchor objects to start the reasoning chain.
    \item \textbf{Construct the Chain:} Create a logical path (\eg, Landmark $\to$ Spatial Relation $\to$ Intermediate Object $\to$ Target).
    \item \textbf{Implicit Inference:} Use scene context (\eg, ``opposite the sink'') rather than just explicit \texttt{relational\_location} fields.
    \item \textbf{MANDATORY:} Include \texttt{visibility\_segments} for ALL referenced objects.
\end{itemize}

\vspace{0.15cm}
\textbf{2.\ CRITICAL CONSTRAINTS \& ANTI-PATTERNS:}
\begin{itemize}[leftmargin=*, nosep]
    \item \textbf{Minimal Descriptions:} Use generic terms (``the machine'') instead of specific attributes (``the Samsung washer'') to force spatial reasoning.
    \item \textbf{Avoid Direct Targeting (\texttimes):} Do NOT describe unique attributes that identify the target without spatial logic.
    \item \textbf{Avoid Breaking the Chain (\texttimes):} Do NOT explicitly state a landmark's absolute location. The location must be found relative to other objects.
\end{itemize}

\vspace{0.15cm}
\textbf{3.\ FEW-SHOT EXAMPLES:}

\textbf{Easy (2 Hops):}
\begin{lstlisting}[language=promptJson]
{
  "question_text": "What object is mounted on the wall above the appliance that sits to the right of the entrance?",
  "answer": "the heated towel rail",
  "rationale": ["1. Anchor: entrance (door_001).",
    "2. Appliance to its right: washing_machine_001.",
    "3. Object mounted above it: heated_towel_rail_001."]
}
\end{lstlisting}

\textbf{Medium (3 Hops):}
\begin{lstlisting}[language=promptJson]
{
  "question_text": "What item is sitting on the edge of the fixture located next to the wall-mounted toilet?",
  "answer": "the bath toy",
  "rationale": ["1. Anchor: toilet_001.",
    "2. Fixture next to it: bathtub_001.",
    "3. Target on the edge: bath_toy_002."]
}
\end{lstlisting}

\textbf{Hard (Inferred Layout \& Virtual Path):}
\begin{lstlisting}[language=promptJson]
{
  "question_text": "Begin at the framed poster in the hallway. If you pass through the nearby door, what furniture has a covered container underneath it?",
  "answer": "the sink vanity",
  "rationale": ["1. Anchor: movie_poster_002 (Hallway).",
    "2. Virtual path: Pass through nearby door_001.",
    "3. Intermediate: cat_litter_box_001 (covered container).",
    "4. Relation: It is under the sink_vanity_001."]
}
\end{lstlisting}
\end{promptbox}
\vspace{-0.2cm}
\caption{System prompt for generating Cross-View Multi-hop Path Reasoning tasks.}
\label{fig:spatialreasoningtemplate}
\end{figure*}


\begin{figure*}[p]
\centering
\begin{promptbox}{Task Template 3: Layout Inference}
\vspace{0.2cm}
\textbf{ROLE:} Generate spatial layout questions based on triplet relationships from a video annotation.

\textbf{INPUT:} Layout Triplets: \texttt{\{\{LAYOUT\_TRIPLETS\}\}}; Object Inventory: \texttt{\{\{OBJECT\_INVENTORY\}\}}. Each triplet describes a spatial relationship $(Object_1, Object_2, Object_3)$ where $Object_2$ acts as an obstacle between $Object_1$ and $Object_3$.

\vspace{0.15cm}
\textbf{1.\ QUESTION CATEGORIES (Difficulty 1--3):}
\begin{enumerate}[leftmargin=*, nosep, label=\textbf{\arabic*.}]
    \item \textbf{Direct Path:} Test immediate accessibility (\eg, ``Can you walk directly from the sink to the toilet?'').
    \item \textbf{Obstacle Identification:} Identify the blocker (\eg, ``What object blocks the direct path from the trash can to the bathtub?'').
    \item \textbf{Alternative Path:} Path planning (\eg, ``If you want to go from the DVD case to the trash can, what do you need to go around?'').
    \item \textbf{Multi-step Navigation:} Complex routing (\eg, ``To reach the door from the sink, which objects would you need to navigate around?'').
    \item \textbf{Spatial Positioning:} Relative location logic (\eg, ``Which object is positioned between the toilet and the towel rack?'').
\end{enumerate}

\vspace{0.15cm}
\textbf{2.\ MANDATORY CONSTRAINTS:}
\begin{itemize}[leftmargin=*, nosep]
    \item \textbf{Data Grounding:} Answers must be logically derived strictly from the provided triplet relationships.
    \item \textbf{Visibility Data:} For EVERY question, you \textbf{MUST} include the \texttt{visibility\_segments} for all referenced objects.
    \item \textbf{Naming:} Use object names exactly as they appear in the triplets.
\end{itemize}

\vspace{0.15cm}
\textbf{3.\ REQUIRED JSON OUTPUT FORMAT:}
\begin{lstlisting}[language=promptJson]
[
  {
    "question_id": "[auto-generated]",
    "question_text": "What object blocks the direct path from the trash can to the bathtub?",
    "answer": "The Toilet",
    "rationale": "Based on triplet (Trash Can, Toilet, Bathtub), the Toilet is the obstacle.",
    "question_type": "layout-reasoning",
    "sub_question_type": "obstacle_identification",
    "level": 2,
    "visibility_segments": [
      {"object_id": "trash_can_001", "visibility_segments": [[10.5, 15.2], [20.1, 25.0]]},
      {"object_id": "toilet_001", "visibility_segments": [[0.0, 30.0]]},
      {"object_id": "bathtub_001", "visibility_segments": [[5.0, 20.0]]}
    ],
    "generated_by": "llm-creative"
  }
]
\end{lstlisting}
\end{promptbox}
\vspace{-0.2cm}
\caption{System prompt for generating Layout Inference tasks.}
\label{fig:layouttemplate}
\end{figure*}


\begin{figure*}[p]
\centering
\begin{promptbox}{Task Template 4: Functional Association}
\vspace{0.2cm}
\textbf{ROLE:} You are an AI assistant specializing in VQA dataset creation. Your task is to generate natural, high-quality questions focusing on \textbf{multi-object functional relationships}.

\textbf{OBJECTIVE:} Create questions that require understanding active functional interactions (data, power, control) between at least 2 objects. Questions must be impossible to answer without video understanding.

\vspace{0.15cm}
\textbf{1.\ CRITICAL RULES:}
\begin{enumerate}[leftmargin=*, nosep, label=\textbf{\arabic*.}]
    \item \textbf{Multi-Object Functional Only:} Focus EXCLUSIVELY on active relationships (data processing, signal transmission, power supply). \textbf{NEVER} use trivial spatial relations like ``support'', ``rests\_on'', or ``near''.
    \item \textbf{Physical Descriptions Only:} Reference objects ONLY by neutral attributes (color, shape, material). \textbf{NEVER} mention function or purpose (avoid ``display'', ``storage'', ``cooking'').
    \item \textbf{Minimal Target Identification:} When asking about the target, use generic terms (``What device...'').
    \item \textbf{Video-Dependent:} Questions must require understanding \textit{interaction}, not just appearance.
\end{enumerate}

\vspace{0.15cm}
\textbf{2.\ QUESTION TYPES \& EXAMPLES:}

\textbf{Level 1 -- Object Processing/Control:}
\begin{itemize}[leftmargin=*, nosep]
    \item \textit{Ask which object processes input from or controls another.}
    \item \textbf{Q:} ``What receives input from that grey and black device on the desk?'' \\ \textbf{A:} ``The black computer tower on the floor.'' \\ \textit{Rationale: mouse (grey/black) $\to$ functionality\_relation $\to$ PC tower.}
\end{itemize}

\textbf{Level 2 -- Spatial-Functional Context:}
\begin{itemize}[leftmargin=*, nosep]
    \item \textit{Ask about objects based on their functional ecosystem.}
    \item \textbf{Q:} ``What device connects to those black rectangular objects and that black keyboard?'' \\ \textbf{A:} ``The black computer case on the floor.''
\end{itemize}

\vspace{0.15cm}
\textbf{3.\ REQUIRED OUTPUT FORMAT:}
\begin{lstlisting}[language=promptJson]
[
  {
    "question_id": "scene_id_XX",
    "level": <1, 2, or 3>,
    "question_type": "Multi_Object_Relationship",
    "question_text": "What object works with that blue fabric item at the desk?",
    "answer": "The light wood cabinet with silver handles under the desk",
    "rationale": "Identified office_chair_001 via functionality_relation to filing_cabinet_001.",
    "objects_involved": [
      {"instance_id": "office_chair_001", "most_prominent_timestamp": "00:01"},
      {"instance_id": "filing_cabinet_001", "most_prominent_timestamp": "00:02"}
    ],
    "generated_by": "llm-creative"
  }
]
\end{lstlisting}

\vspace{0.15cm}
\textbf{KEY REQUIREMENTS:}
\begin{itemize}[leftmargin=*, nosep]
    \item Use \texttt{functionality\_relation}, \texttt{relational\_location}, and \texttt{functionalEcosystem} data extensively.
    \item Focus on: control, processing, data transmission, signal flow, power supply.
\end{itemize}
\end{promptbox}
\vspace{-0.2cm}
\caption{System prompt for generating Functional Association tasks.}
\label{fig:functionalasscociationtemplate}
\end{figure*}


\end{document}